\documentclass{article}

\usepackage{arxiv}

\usepackage[utf8]{inputenc} 
\usepackage[T1]{fontenc}    
\usepackage{hyperref}       
\usepackage{url}            
\usepackage{booktabs}       
\usepackage{amsfonts}       
\usepackage{nicefrac}       
\usepackage{microtype}      
\usepackage{graphicx}
\usepackage{natbib}
\usepackage{doi}
\usepackage{amsmath} 
\usepackage{caption}

\title{Contrastive Representation Learning Helps Cross-institutional Knowledge Transfer: \\A Study in Pediatric Ventilation Management}
\date{}

\author{
\begin{center}
\begin{minipage}{0.95\textwidth}
\begin{tabular*}{\textwidth}{@{\extracolsep{\fill}}cc}
    \\
    \begin{minipage}[t]{0.5\textwidth}
        \centering
        \href{https://orcid.org/0009-0002-2586-6479}{\includegraphics[scale=0.06]{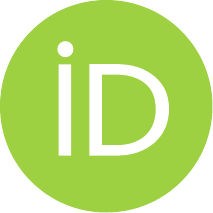}\hspace{1mm}Yuxuan Liu} \\
        \mdseries UKRI Centre for Doctoral Training in AI for Healthcare\\
        \mdseries Department of Computing\\
        \mdseries Imperial College London, UK \\
        \texttt{edison.liu22@imperial.ac.uk}
    \end{minipage}
    &
    \begin{minipage}[t]{0.5\textwidth}
        \centering
        \href{https://orcid.org/0000-0003-0156-5065}{\includegraphics[scale=0.06]{orcid.pdf}\hspace{1mm}Jinpei Han} \\
        \mdseries Department of Computing\\
        \mdseries Imperial College London, UK \\
        \texttt{j.han20@imperial.ac.uk}
    \end{minipage}
    \\[5em]
    \begin{minipage}[t]{0.5\textwidth}
        \centering
        \href{https://orcid.org/0000-0003-0784-8154}{\includegraphics[scale=0.06]{orcid.pdf}\hspace{1mm}Padmanabhan Ramnarayan} \\
        \mdseries Department of Surgery \& Cancer\\
        \mdseries Imperial College London, UK \\
        \texttt{p.ramnarayan@imperial.ac.uk}
    \end{minipage}
    &
    \begin{minipage}[t]{0.5\textwidth}
        \centering
        \href{https://orcid.org/0000-0003-0813-7207}{\includegraphics[scale=0.06]{orcid.pdf}\hspace{1mm}A. Aldo Faisal} \\
        \mdseries Department of Computing\\
        \mdseries Imperial College London, UK \\
        \mdseries Chair in Digital Health \& Data Science\\
        \mdseries University of Bayreuth, Germany \\
        \texttt{aldo.faisal@imperial.ac.uk}
    \end{minipage}
\end{tabular*}
\end{minipage}
\end{center}
}

\renewcommand{\headeright}{}
\renewcommand{\undertitle}{Accepted for presentation at the AAAI'25 Workshop on Health Intelligence}
\renewcommand{\shorttitle}{Contrastive Representation Learning for Pediatric Extubation Readiness Assessment}

\hypersetup{
pdftitle={Contrastive Representation Learning Helps Cross-institutional Knowledge Transfer: A Study in Pediatric Ventilation Management},
pdfsubject={cs.LG, cs.AI, q-bio.QM},
pdfauthor={Yuxuan Liu, Jinpei Han, Padmanabhan Ramnarayan, A. Aldo Faisal},
pdfkeywords={Contrastive Learning, Clinical Decision Making, Knowledge Transfer, Pediatric Ventilation},
}

\begin{document}
\maketitle

\begin{abstract}
Clinical machine learning deployment across institutions faces significant challenges when patient populations and clinical practices differ substantially. We present a systematic framework for cross-institutional knowledge transfer in clinical time series, demonstrated through pediatric ventilation management between a general pediatric intensive care unit (PICU) and a cardiac-focused unit. Using contrastive predictive coding (CPC) for representation learning, we investigate how different data regimes and fine-tuning strategies affect knowledge transfer across institutional boundaries. Our results show that while direct model transfer performs poorly, CPC with appropriate fine-tuning enables effective knowledge sharing between institutions, with benefits particularly evident in limited data scenarios. Analysis of transfer patterns reveals an important asymmetry: temporal progression patterns transfer more readily than point-of-care decisions, suggesting practical pathways for cross-institutional deployment. Through a systematic evaluation of fine-tuning approaches and transfer patterns, our work provides insights for developing more generalizable clinical decision support systems while enabling smaller specialized units to leverage knowledge from larger centers.
\end{abstract}


\section{Introduction}
\label{sec:1}

\begin{figure}[!t]
    \centering
    \includegraphics[width=0.65\textwidth]{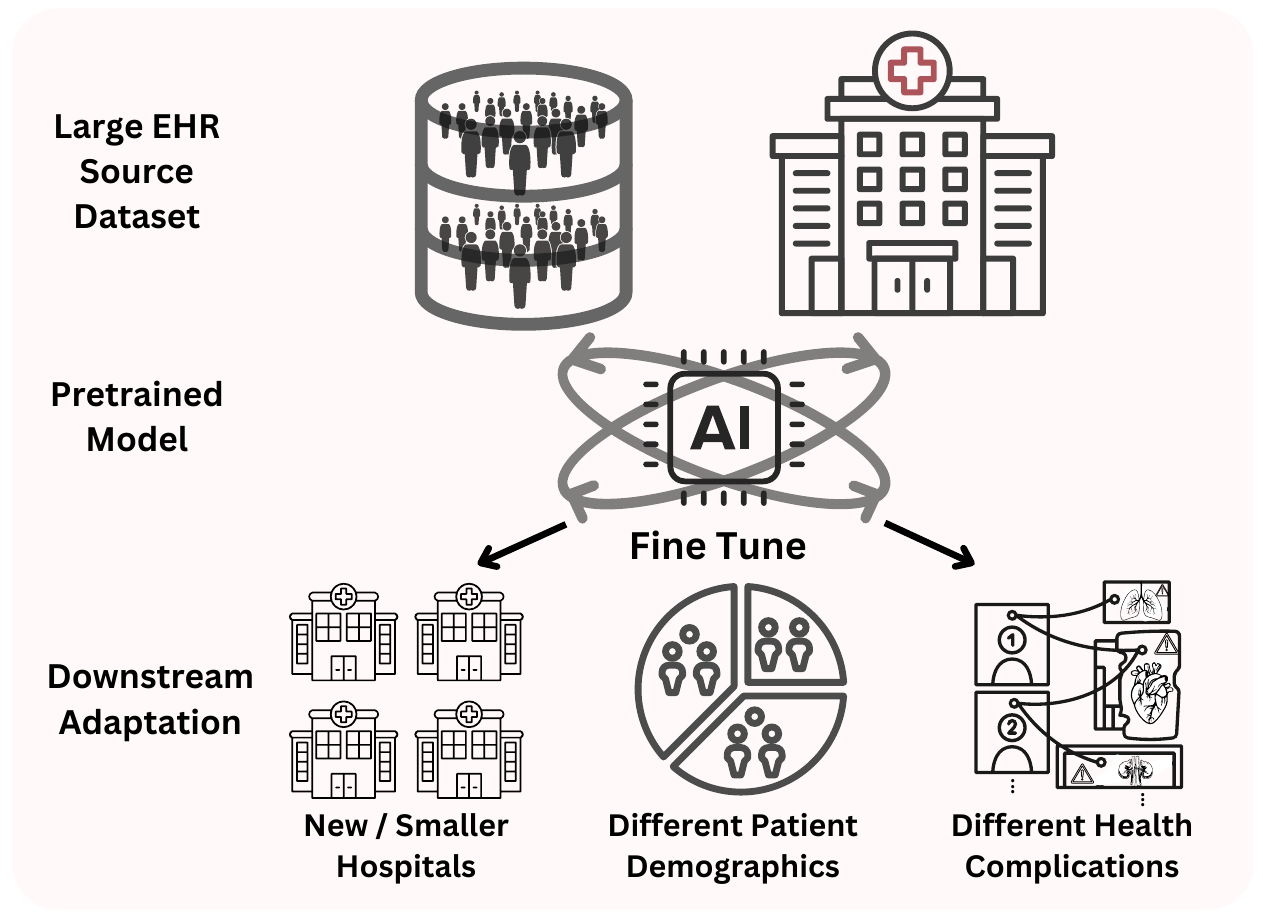}
    \caption{Overview of Cross-institutional Knowledge Transfer in Healthcare: From Large Source Data to Specialized Adaptation.}
    \label{fig:concept}
    \vspace{-0.5em}
\end{figure}

Machine learning has shown promising results in clinical decision support, particularly for complex intensive care settings \citep{gottesman2019guidelines}. However, developing robust models faces significant challenges: limited data availability, variations in clinical practices across institutions, and restricted data sharing. These constraints often result in models that perform well locally but fail to generalize across different clinical settings \citep{mcdermott2021reproducibility}. This cross-site generalization problem represents a fundamental challenge in the real-world application of clinical ML, particularly when dealing with longitudinal patient data in Electronic Healthcare Records (EHR).

Recent advances in generative AI and large foundation models have demonstrated the power of self-supervised representation learning in capturing transferable features from unlabeled data \citep{bommasani2021FM,brown2020language}. This capacity is particularly valuable for EHR applications, where obtaining high-quality labeled data is both costly and resource-intensive. Despite growing interest and successful applications of self-supervised learning to EHR time series data \citep{rasmy2021med, tu2024biogeneralist, wornow2023ehrshot}, downstream evaluations have largely been restricted to single-institution settings, where test data, though held out, still originates from the same underlying population as the training data. This leaves open the critical question of how representation learning can facilitate knowledge transfer between institutions with different patient populations and clinical practices \citep{mcdermott2021comprehensive}.

In this work, we investigate cross-institutional knowledge transfer in the context of pediatric mechanical ventilation, focusing on the challenging scenario of sharing clinical expertise between a large general PICU and a smaller cardiac-focused unit (Fig. \ref{fig:concept}). This scenario presents an ideal testbed as ventilation management directly impacts patient outcomes—both prolonged ventilation and premature extubation increase complications and costs \citep{adverseEvents, CostAnalysis}—yet achieving standardization remains difficult due to rapid physiological changes in pediatric patients and varying institutional practices. To derive transferable representations in this challenging setting, we adopt contrastive learning as our self-supervised learning approach. This group of methods align naturally with clinical reasoning by embedding physiologically related states close together while separating unrelated ones \citep{liu2023self}. Our contributions are:

\begin{itemize}
    \item A systematic framework for studying cross-institutional knowledge transfer in clinical time series, with comprehensive evaluation strategies across different data regimes and fine-tuning strategies
    \item Empirical evidence that contrastive learning enables effective knowledge sharing between institutions with distinct patient populations, demonstrated through pediatric ventilation management
\end{itemize}

Our results demonstrate that while performance gaps exist between institutions, contrastive pre-training with appropriate fine-tuning strategies can significantly reduce these gaps, particularly in few-shot learning scenarios. More specifically, our analysis reveals asymmetric transfer patterns across different clinical tasks, suggesting practical pathways for cross-institutional deployment of clinical ML systems. These findings contribute to the broader understanding of how representation learning can facilitate knowledge sharing between institutions with distinct patient populations and clinical practices.

\section{Methods}
\label{sec:2}
\begin{figure*}[!t]
    \centering
    \includegraphics[width=\textwidth]{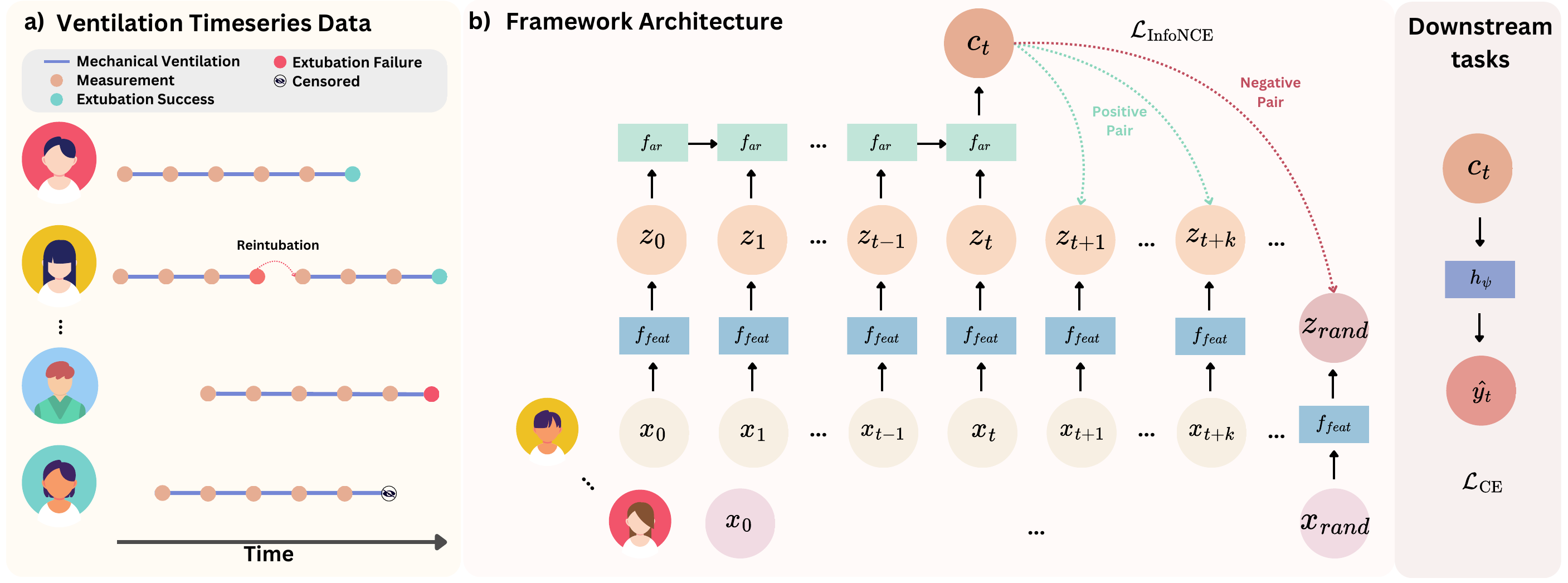}
    \caption{\textbf{(A) Ventilation Time Series Data Types:} Our dataset contains successful extubation, reintubation, failed extubation and censored data with unobservable outcomes. \textbf{(B) Framework Architecture:} Network $f_\theta$ Pre-training on source institution data using modified CPC, then fine-tunned with classification layer $h_\psi$ in downstream tasks.}
    \label{fig:pipeline}
\end{figure*}

This section presents a framework for cross-institutional knowledge transfer in clinical time series analysis. While demonstrated through pediatric ventilation management, our methodology generalizes to other clinical scenarios where institutional differences create natural barriers to model generalization.

\subsection{Problem Formulation}
\label{subsec:2.1}
Let $\mathcal{D}_s = \{\mathbf{x}_i^s, \mathbf{y}_i^s\}_{i=1}^{N_s}$ denote the source domain dataset with $N_s$ samples, where each $\mathbf{x}_i^s \in \mathbb{R}^{T \times D}$ represents a multivariate time series with $T$ timesteps and $D$ features, and $\mathbf{y}_i^s$ represents the corresponding labels. Similarly, let $\mathcal{D}_t = \{\mathbf{x}_i^t, \mathbf{y}_i^t\}_{i=1}^{N_t}$ denote the target domain dataset.

Our framework implements cross-institutional knowledge transfer through two phases (Fig. \ref{fig:pipeline} (B)): representation learning on source institution data (pre-training) and model adaptation for the target institution (fine-tuning). For pre-training, we learn a representation function $f_\theta: \mathbb{R}^{T \times D} \rightarrow \mathbb{R}^{H}$ that maps input time series to a $H$-dimensional latent space, where $\theta$ denotes the parameters of the encoder network. Following the setup in \citep{mcdermott2021comprehensive}, in the context of transfer learning between institutions, we consider three approaches for learning predictive models:

\begin{equation}
\begin{array}{ll@{\qquad}l}
    \text{Target-Only:} & \hat{y} = h_\psi(g_\phi(\mathbf{x})), & \{\phi, \psi\} \subset \Theta \\[0.5ex]
    \text{FTF:} & \hat{y} = h_\psi(f_\theta(\mathbf{x})), & \{\theta, \psi\} \subset \Theta \\[0.5ex]
    \text{FTD:} & \hat{y} = h_\psi(f_\theta(\mathbf{x})), & \{\psi\} \subset \Theta, \theta = \theta_s
\end{array}
\end{equation}
where $g_\phi$ represents the target-trained encoder, $h_\psi$ denotes a task-specific decoder with parameters $\psi$, and $f_\theta$ is the source-pretrained encoder. $\Theta$ represents the set of trainable parameters, and $\theta_s$ denotes the fixed source-pretrained parameters. The target-only approach trains both components from scratch. In contrast, both FTF (Fine-Tuning Full) and FTD (Fine-Tuning Decoder-only) leverage the source-pretrained encoder: FTF updates all parameters during training, while FTD keeps the encoder fixed and only updates decoder parameters.

\subsection{Study Design}
\label{subsec:2.2}
\subsubsection{Clinical Setting and Data Preparation}
We analyze ventilation data from 2013-2022 across two PICUs: a general unit (1,883 episodes), and a cardiac-focused unit (1,932 episodes, hereafter the target institution for our transfer learning approach). The institutions differ substantially in patient demographics and clinical characteristics: the target institution treats predominantly cardiac patients (84.8\% vs 1.6\%), serves younger patients (median age 5.0 vs 16.0 months), requires more intensive intervention (vasoactive support: 55.5\% vs 20.4\%), and has shorter ventilation durations (48.0 vs 80.0 hours). These distinct clinical characteristics serve as an exemplar case for investigating cross-institutional transfer learning methods.

For each ventilation episode, we collect 37 hourly-sampled clinical variables spanning demographics, vital signs, ventilator parameters, laboratory values, and medications (see Appendix Table \ref{appendix:featurelist}). Inclusion criteria required episodes using conventional pressure control/support modes, lasting between 12 hours and 28 days. To maintain statistical independence between samples, we analyze only the first extubation attempt for each admission, as outcomes of subsequent attempts are inherently influenced by clinical decisions and physiological responses from previous attempts. Censored cases (166 episodes) where extubation outcomes were unknown due to transfer out or death were excluded from supervised training but utilized during contrastive pre-training (see Fig. \ref{fig:pipeline} (A)).

Following the data preprocessing framework established in \citep{wang2020mimic}, we aggregate data into hourly timestamps, with continuous variables (e.g. vital signs, laboratory values) represented by their median values within each hour window and categorical variables (e.g. ventilation modes, medication types) by their most frequent occurrence. Missing data were handled through a two-step approach to ensure complete cases for analysis: values were first forward-filled up to 12 hours following clinical assessment patterns, then completed using multivariable nearest-neighbour imputation with physiologically related predictors.

\begin{table}[!t]
\caption{Patient Demographics and Clinical Characteristics}
\label{tab:demographics}
\begin{tabular*}{\textwidth}{@{\extracolsep{\fill}}lll@{}}
\hline
Characteristic & Source Inst. & Target Inst.\\
\hline
\multicolumn{3}{l}{\textbf{Demographics}} \\[0.5ex]
Unique ventilation episodes (N) & 1,883 & 1,932 \\
Age, months (Median, IQR)\textsuperscript{†} & 16.0 (3.0-62.5) & 5.0 (1.0-14.0) \\
Male gender (N, \%) & 1,077 (57.2) & 1,068 (55.3) \\[1ex]
\multicolumn{3}{l}{\textbf{Clinical Characteristics}} \\[0.5ex]
Primary diagnosis (N, \%) & & \\
\ \ Cardiovascular & 29 (1.6) & 1,638 (84.8) \\
\ \ Respiratory & 995 (52.8) & 129 (6.7) \\
\ \ Others & 859 (45.6) & 165 (8.5) \\
PIM  (Median, IQR)\textsuperscript{†} & 0.014 (0.006-0.053) & 0.022 (0.011-0.053)\\
Vasoactive support (N, \%) & 385 (20.4) & 1,072 (55.5) \\
Length of Episode, hours (Median, IQR)\textsuperscript{†} & 80.0 (46.0-124.0) & 48.0 (23.0-111.0) \\
\hline
\multicolumn{3}{l}{\textsuperscript{†}IQR: Interquartile Range}\\
\end{tabular*}
\end{table}

\subsubsection{Prediction Tasks}
We investigate two complementary ventilation weaning tasks that test different aspects of knowledge transfer:

\begin{enumerate}
\item \textbf{Point-of-care Extubation Risk Assessment:} At the time of planned ventilator removal, this task predicts the probability of extubation failure, defined as requiring endotracheal tube reinsertion within 48 hours post-extubation \citep{thille2011outcomes}. Extubation failure occurs in approximately 10\% of cases and represents a preventable adverse event requiring emergency intervention.

\item \textbf{Prospective Extubation Window Identification:} This task continuously assesses the probability of successful extubation over the subsequent 12-hour window. These windows, occurring in roughly 15\% of monitoring periods, identify the optimal timing for clinical evaluation of ventilation liberation.

\end{enumerate}
The point-of-care assessment primarily challenges a model's ability to adapt to institution-specific risk thresholds, while window identification tests the transfer of learned temporal dynamics across different patient populations. Both tasks predict minority class events following standard statistical practice.

\subsection{Pre-training with Contrastive Learning}

Contrastive learning has been widely explored in clinical time series analysis, with various approaches targeting different aspects of temporal data modeling \citep{krishnan2022self, liu2023self}. Early methods such as Temporal Neighborhood Coding (TNC) \citep{tonekaboni2021unsupervised} and SOM-CPC \citep{huijben2023som} demonstrated success in learning representations from sensor-based data, including physiological signals (ECG, EEG) and human activity recordings. These data types typically feature consistent sampling rates and clear temporal patterns: TNC leverages this temporal regularity to capture progression patterns, while SOM-CPC combines contrastive learning with self-organizing maps for interpretable visualization.

EHR time series present additional challenges beyond sensor data, including irregular sampling, missing values, and complex interdependencies between variables. Contrastive Predictive Coding (CPC) \citep{oord2018representation}, which constructs representations by predicting future states while discriminating against unrelated samples, offers a natural framework for handling these challenges. Recent work \citep{bouchattaoui2024causal} has demonstrated CPC's effectiveness on EHR data through its application in causal inference tasks. Given this proven adaptability, we adopt the standard CPC framework for our cross-institutional knowledge transfer investigation, providing an effective balance between capturing temporal dependencies and computational efficiency.

\subsubsection{CPC Architecture}
Given an input sequence $\mathbf{x}^s \in \mathbb{R}^{T \times D}$ from the source domain, let $\mathbf{x}_t^s \in \mathbb{R}^D$ denote its features at time step $t$. The CPC framework consists of two main components (omitting sample index $i$ for clarity): a feature extractor $f_{\text{feat}}$ that maps each timestep to an intermediate representation $\mathbf{z}_t^s = f_{\text{feat}}(\mathbf{x}_t^s)$, and an autoregressive model $f_{\text{ar}}$ that aggregates these representations $\mathbf{c}_t^s = f_{\text{ar}}(\mathbf{z}_{\leq t}^s)$. Together, these components ($f_{\text{feat}}$ and $f_{\text{ar}}$) form the representation function $f_\theta$ referenced in our problem formulation. Following \citep{oord2018representation}, we optimize the InfoNCE loss that aims to predict K future representations while contrasting against negative samples:

\begin{eqnarray}
   \mathcal{L}_{\text{InfoNCE}} = -\mathbb{E}_{\mathbf{x}^s \sim \mathcal{D}_s}\left[\sum_{k=1}^K \log \frac{\exp(T_k(\mathbf{c}_t^s, \mathbf{z}_{t+k}^s))}{\sum_{\mathbf{z}' \in \mathcal{N}} \exp(T_k(\mathbf{c}_t^s, \mathbf{z}'))}\right]
\end{eqnarray}
where $T_k(\cdot, \cdot)$ is a step-specific parameterized discriminator that measures compatibility between representations for k-step ahead prediction, and $\mathcal{N}$ contains negative samples drawn from the same batch. After the pre-training, the task-specific decoder $h_\psi$ could be trained by minimizing the Cross-Entropy loss $\mathcal{L}_{\text{CE}}$ on downstream tasks.

\subsubsection{Clinical Domain Adaptation}
Standard CPC employs random negative sampling, which implicitly assumes that all non-positive pairs provide equally valuable contrasting information. However, in EHR time series, this assumption overlooks the inherent structure of patient trajectories - patients with similar clinical characteristics (e.g., age, comorbidities, mortality risk scores) may exhibit more similar temporal patterns. Random sampling might, therefore, select predominantly ``easy'' negative examples, failing to encourage the model to learn fine-grained discriminative features that separate clinically similar cases.

We incorporate this domain knowledge through a simple modification of the sampling strategy. Let $m(\mathbf{x}_{\leq t}^s, \mathbf{x}_{\leq t}'^s) \rightarrow \mathbb{R}$ measure the clinical relevance between two sequences up to time t. The sampling probability for negative pairs is then defined as:

\begin{equation}
p(\mathbf{z}' \mid \mathbf{z}_t^s) \propto \exp(\beta \cdot m(\mathbf{x}_{\leq t}^s, \mathbf{x}_{\leq t}'^s))
\end{equation}

where $\beta$ is a temperature parameter. This modifies the InfoNCE loss to:

\vspace{-1em}
\begin{eqnarray}
    \mathcal{L}_{\text{InfoNCE}}^{\text{guided}} = -\mathbb{E}_{\mathbf{x}^s \sim \mathcal{D}_s}\left[\sum_{k=1}^K \log \frac{\exp(T_k(\mathbf{c}_t^s, \mathbf{z}_{t+k}^s))}{\sum_{\mathbf{z}' \sim p(\cdot \mid \mathbf{z}_t^s)} \exp(T_k(\mathbf{c}_t^s, \mathbf{z}'))}\right]
\end{eqnarray}
where negative samples are drawn according to our similarity-based distribution rather than uniformly, this targeted sampling helps guide the model in learning representations that capture meaningful clinical variations during contrastive learning.

\subsection{Implementation and Evaluation}
For implementing the CPC framework, we adopt a Multilayer Perceptron (MLP) as feature extractor $f_{\text{feat}}$ and a Gated Recurrent Unit (GRU) as the autoregressive model $f_{\text{ar}}$. The similarity-guided sampling leverages time-to-extubation (TTE) as the proxy measure $m(\cdot,\cdot)$, motivated by the clinical observation that patients at similar stages of respiratory recovery exhibit comparable physiological patterns.

Following \citep{mcdermott2021comprehensive}, we implement a simpler architecture as a baseline during direct learning, where $f_{\text{feat}}$ is replaced by a linear projection layer while maintaining the same GRU architecture. Both models utilize task-specific linear heads $h_\psi$ for downstream prediction.

All hyperparameters were tuned using grid search on the respective validation sets: source domain validation set for pre-training and target domain validation set for fine-tuning. Detailed configurations are provided in Appendix Table \ref{appendix:hyperparameters}.

We evaluate three approaches as defined in Sect.~\ref{subsec:2.1}: direct training on target domain (Target-Only), full model fine-tuning (FTF), and decoder-only fine-tuning with fixed pre-trained representations (FTD). For the ablation study, we evaluated both CPC and GRU variants as the encoder network $f_\theta$. To simulate realistic scenarios of limited data availability, we conduct few-shot experiments using 30\% and 5\% of the target domain data during training. All experiments maintain a fixed train/validation/test split of 65\%/15\%/20\%. For each experimental setting, we train models with 5 different random seeds and report the mean and standard deviation.

Performance is assessed using area under the receiver operating characteristic curve (AUROC), area under the precision-recall curve (AUPRC), and balanced accuracy, providing comprehensive evaluation across different operating points given the class imbalance nature of our tasks.

\section{Results \& Discussion}
\label{sec:3}

\begin{table}[!t]
\caption{Model Performance on Target Institution Under Different Data Regimes}
\label{tab:few_shot}
\setlength{\tabcolsep}{6.5pt}  
\renewcommand{\arraystretch}{1.1}
\begin{tabular*}{\textwidth}{@{\extracolsep{\fill}}l|ccc|ccc@{}}
\hline
& \multicolumn{3}{c|}{Task 1} & \multicolumn{3}{c}{Task 2} \\
Model & AUROC & AUPRC & B-Acc & AUROC & AUPRC & B-Acc \\
\hline
\multicolumn{7}{l}{\textit{100\% Training Data}} \\
Source-Only & .709±.028 & .289±.041 & .542±.027 & .801±.013 & .414±.026 & .635±.034 \\
Target-Only & \textbf{.785±.030} & .481±.052 & .714±.037 & \textbf{.857±.011} & \textbf{.509±.031} & \textbf{.767±.007} \\
Source+FTD & .745±.046 & .418±.065 & .691±.048 & .842±.008 & .475±.021 & .740±.008 \\
Source+FTF & .768±.028 & .481±.044 & .721±.032 & .856±.010 & .505±.029 & .765±.009 \\
CPC+FTD & .738±.046 & .393±.042 & .619±.061 & .813±.019 & .423±.026 & .703±.029 \\
CPC+FTF & .788±.034 & \textbf{.510±.066} & \textbf{.716±.017} & .852±.007 & .499±.026 & .765±.009 \\[0.5ex]
\multicolumn{7}{l}{\textit{30\% Training Data}} \\
Target-Only & .755±.015 & .410±.045 & .671±.041 & \textbf{.839±.002} & \textbf{.480±.018} & \textbf{.756±.002} \\
Source+FTD & .741±.046 & .385±.033 & .675±.058 & .830±.005 & .451±.026 & .740±.006 \\
Source+FTF & .756±.033 & .408±.037 & .636±.045 & .836±.009 & .477±.032 & .756±.007 \\
CPC+FTD & .729±.048 & .387±.022 & .631±.078 & .811±.018 & .419±.044 & .709±.016 \\
CPC+FTF & \textbf{.764±.037} & \textbf{.434±.032} & \textbf{.663±.075} & .832±.007 & .455±.017 & .753±.005 \\[0.5ex]
\multicolumn{7}{l}{\textit{5\% Training Data}} \\
Target-Only & .716±.047 & .302±.058 & .611±.049 & .789±.004 & \textbf{.399±.032} & .704±.011 \\
Source+FTD & .661±.085 & .240±.083 & .573±.046 & .772±.016 & .371±.038 & .670±.015 \\
Source+FTF & .708±.045 & .287±.050 & .600±.061 & .785±.013 & .389±.023 & .700±.019 \\
CPC+FTD & .510±.071 & .141±.046 & .500±.002 & .748±.033 & .352±.017 & .599±.022 \\
CPC+FTF & \textbf{.736±.059} & \textbf{.412±.054} & \textbf{.619±.070} & \textbf{.797±.010} & .391±.031 & \textbf{.708±.012} \\
\hline
\end{tabular*}

\vspace{1ex}
\footnotesize
AUROC: Area Under ROC Curve; AUPRC: Area Under Precision-Recall Curve; B-Acc: Balanced Accuracy; FTD: Fine-tune Decoder; FTF: Fine-tune Full model; Source-Only: Direct transfer without fine-tuning
\end{table}
Our experimental results demonstrate significant performance variations across transfer learning strategies and data regimes (Table \ref{tab:few_shot}). Direct application of source institution models to the target domain performs substantially lower than models trained directly on target data (AUROC drops from 0.785 to 0.709 in Task 1 and 0.857 to 0.801 in Task 2, both $p<0.01$). This performance gap highlights a fundamental challenge in clinical ML deployment - the impact of institutional specialization on model generalization. The substantial degradation suggests that differences in patient populations and clinical practices manifest as systematic shifts in physiological patterns rather than simple variations in feature distributions.

The CPC framework with full model fine-tuning (CPC-FTF) effectively bridges this institutional divide, particularly in limited data scenarios. With only 5\% data, CPC-FTF significantly outperforms target-only training in Task 1 (AUROC 0.736 vs 0.716, $p<0.05$) and maintains comparable performance in Task 2 (AUROC 0.797 vs 0.789), while achieving similar results to full data training in both tasks. This pattern suggests that CPC-based pre-training can learn representations that transfer effectively across institutions while allowing for task-specific adaptation.

Analysis of fine-tuning strategies reveals deeper insights into the nature of transferable clinical knowledge. Decoder-only fine-tuning consistently underperforms full model fine-tuning across both tasks and all data regimes, with this gap most pronounced in few-shot learning. When reducing from full to 5\% target data, FTD degrades significantly (Task 1: 0.736 vs 0.510, $p<0.001$; Task 2: 0.797 vs 0.748, $p<0.05$), while FTF maintains relatively robust performance. This suggests that effective transfer requires both preserving general physiological patterns and adapting feature extractors to institution-specific variations, challenging the common practice of using pre-trained models as fixed feature extractors.

Task-specific analysis reveals an important asymmetry in knowledge transfer that has implications for deployment strategy. Task 2 (continuous window identification) shows consistently higher baseline performance and smaller transfer learning gaps compared to Task 1 (point-of-care prediction). This difference suggests that models more readily transfer knowledge of temporal progression patterns, which benefit from extended monitoring windows and richer contextual information. In contrast, point-of-care decisions appear more sensitive to institution-specific factors, likely due to their dependence on precise threshold judgments that vary between cardiac and general units. This indicates a practical deployment strategy: beginning with continuous monitoring tasks to establish baseline transferability before tackling more sensitive point-of-care decisions.

Our findings have important implications for clinical ML deployment while highlighting key areas for future work. While we demonstrate effective transfer learning between distinct PICUs, validation across more diverse clinical settings would better establish generalizability. Though CPC effectively learns transferable representations, two critical questions remain: understanding which physiological patterns transfer successfully through systematic model analysis and interpretability studies, and comparing CPC with other self-supervised approaches (e.g., generative-based or adversarial-based approaches \citep{zhang2024self}) to identify optimal architectures for representation learning in clinical time series. These investigations would advance our understanding of transferable clinical knowledge while informing the design of more robust cross-institutional learning systems.

\section{Conclusion}
\label{sec:4}
This work establishes a systematic framework for cross-institutional knowledge transfer in clinical time series, demonstrated through pediatric ventilation management. Our results show that while direct model transfer is ineffective, contrastive pre-training with appropriate fine-tuning enables robust knowledge sharing between institutions with distinct patient populations. The observed asymmetry in transfer success across prediction tasks - with temporal progression patterns transferring more readily than point-of-care decisions - provides important guidance for deploying clinical decision support tools across institutions. Through systematic evaluation of fine-tuning strategies and transfer patterns, our work contributes to the broader goal of enabling reliable knowledge sharing across healthcare institutions while maintaining their clinical autonomy.

\section*{Acknowledgments}  
\addcontentsline{toc}{section}{Acknowledgments}
Y. Liu is supported by UK Research and Innovation UKRI Centre for Doctoral Training in AI for Healthcare grant number EP/S023283/1; P. Ramnarayan is supported by the Imperial College Biomedical Research Centre and the National Institute of Health Research. A. A Faisal is supported by the UKRI Turing AI Fellowship grant number EP/V025449/1.
\clearpage  

\section*{Appendix}
\addcontentsline{toc}{section}{Appendix}

\begin{table*}[!h]
\captionsetup{justification=raggedright, singlelinecheck=false}
\caption{Model Hyperparameters}
\label{appendix:hyperparameters}
\begin{tabular}{llr}
\hline\noalign{\smallskip}
\textbf{Component} & \textbf{Parameter} & \textbf{Range} \\
\noalign{\smallskip}\hline\noalign{\smallskip}
Neural Network & Type & MLP/GRU \\
Architecture & Window size & Int[1, 48] \\
& Layers & Choice[1, 2, 3] \\ 
& Hidden dim & Choice[32, 64, 128] \\
& Dropout & Uniform[0.1, 0.5] \\
& Activation & LeakyReLU \\
& Bidirectional & Choice[True, False] \\
\noalign{\smallskip}\hline\noalign{\smallskip}
Contrastive & K & 4 \\
Learning & \# Pos samples & Choice[2, 4, 8] \\
& \# Neg sample per pos & Choice[4, 8] \\
& Temperature & Uniform[1.0, 5.0] \\
\noalign{\smallskip}\hline\noalign{\smallskip}
Training & Learning rate & Log-uniform[1e-4, 1e-3] \\
Parameters & Weight decay & Log-uniform[1e-5, 1e-4] \\
& Batch size & 128 \\
& Max epochs & 100 \\
& Early stop & Choice[5, 10] \\
& Focal loss $\alpha$ & Uniform[0.5, 1.0] \\
& Focal loss $\gamma$ & Choice[2, 3, 4] \\
\noalign{\smallskip}\hline
\end{tabular}
\end{table*}

\begin{table*}[!h]

\captionsetup{justification=raggedright, singlelinecheck=false}
\caption{Clinical Features Used in Model Development}
\label{appendix:featurelist}
\begin{tabular}{l|p{0.75\textwidth}}
\hline\noalign{\smallskip}
\textbf{Category} & \multicolumn{1}{c}{\textbf{Features}} \\
\noalign{\smallskip}\hline\noalign{\smallskip}
Patient Characteristics & Age, Gender, Weight, PIM-3 score, Diagnosis (Cardiovascular, Respiratory, Neurological, Gastrointestinal, Infection, Others) \\
Vital Signs & Heart Rate, Temperature, SpO2, Blood Pressure (Systolic, Mean, Diastolic) \\
Ventilator Parameters & Mode (PC, PS), PIP, PEEP (set), Respiratory Rate (set), FiO2, Inspiratory Time, Pressure Support \\
Respiratory Measurements & End-tidal CO2, Respiratory Rate (measured), Tidal Volume, Mean Airway Pressure \\
Blood Gas & pH, PCO2, Base Excess, Lactate, HCO3 \\
Laboratory Values & WBC, Neutrophil count, Hemoglobin \\
Medications & NMB, Sedation, Furosemide, Vasoactive agents, Steroids \\
Other & Fluid Balance, Ventilator Hours \\
\noalign{\smallskip}\hline
\end{tabular}
\end{table*}
\clearpage  

\setcitestyle{numbers}
\bibliographystyle{unsrtnat}  
\bibliography{references}

\end{document}